\documentclass[letterpaper, 10 pt, conference]{ieeeconf}
\IEEEoverridecommandlockouts  
\usepackage{amsmath,amsfonts}
\usepackage{algorithmic}
\usepackage{algorithm}
\usepackage{array}
\usepackage{textcomp}
\usepackage{stfloats}
\usepackage{url}
\usepackage{verbatim}
\usepackage{graphicx}
\usepackage{cite}
\usepackage[breaklinks,hidelinks]{hyperref}
\usepackage{cleveref}
\usepackage{siunitx}
\usepackage[nolist,nohyperlinks]{acronym}
\usepackage{booktabs}
\usepackage{makecell}
\usepackage{multirow}

\acrodef{hil}[HIL]{Hardware-In-the-Loop}
\acrodef{rl}[RL]{Reinforcement Learning}
\acrodef{mbrl}[MBRL]{Model-Based Reinforcement Learning}
\acrodef{ssl}[SSL]{Self-Supervised Learning}
\acrodef{ou}[OU]{Ornstein-Uhlenbeck}
\acrodef{mlp}[MLP]{Multi-Layer Perceptron}
\acrodef{lstm}[LSTM]{Long Short-Term Memory}
\acrodef{fov}[FOV]{Field-of-View}
\acrodef{appo}[APPO]{Asynchronous Proximal Policy Optimization}
\acrodef{rnn}[RNN]{Recurrent Neural Network}
\acrodef{gru}[GRU]{Gated Recurrent Unit}
\acrodef{pomdp}[POMDP]{Partially Observable Markov Decision Process}
\acrodef{jepa}[JEPA]{Joint Embedding Predictive Architectures}
\acrodef{rssm}[RSSM]{Recurrent State-Space Model}
\acrodef{kl}[KL]{Kullback-Leibler}
\acrodef{ema}[EMA]{Exponential Moving Average}
\acrodef{ssim}[SSIM]{Structural Similarity Index Measure}
\acrodef{mse}[MSE]{Mean Squared Error}
\acrodef{ood}[OOD]{Out-Of-Distribution}
\acrodef{ppo}[PPO]{Proximal Policy Optimization}
\acrodef{imu}[IMU]{Inertial Measurement Unit}

\title{\LARGE \bf Flying on Imagination: DreamerV3-Style World Models under Environment Distribution Shift for Quadrotor Navigation}

\author{Luca Zanatta$^{*}$, Grzegorz Malczyk$^{*}$, and Kostas Alexis%
\thanks{$^*$ The authors contributed equally.}
\thanks{This work was supported by the Horizon Europe Grant No. 101120732 and the Research Council of Norway projects under Award NO-338694, NO-357451. The authors are with the Department of Engineering Cybernetics, Norwegian University of Science and Technology (NTNU), Norway.}
\thanks{Corresponding author: \tt\small grzegorz.malczyk@ntnu.no}
}

\begin{document}

\maketitle
\thispagestyle{empty}
\pagestyle{empty}

\begin{abstract}

DreamerV3-style reconstructive world models offer a route to vision-based navigation for aerial robots through learned latent dynamics. Yet their behavior under shifts in the training environment and their ability to support navigation using imagined observations, predictions generated by rolling the latent dynamics forward without new sensory input, remain insufficiently characterized. We study four training-environment regimes for depth-based quadrotor navigation, spanning a fixed layout, a finite set of layouts, low-discrepancy continuous obstacle sampling, and i.i.d. uniform sampling. We cross-evaluate reconstruction before and after Reinforcement Learning fine-tuning, separately evaluate the resulting policies in simulation, and transfer them to a real quadrotor without real-world fine-tuning. We demonstrate onboard flight driven by imagined state and depth: a \SI{24.7}{\meter} mission in which $66\%$ of flight time and $76\%$ of traveled distance occur during repeated \SI{5}{\second} sensor-free imagination windows, with prediction quality degrading over each open-loop interval and recovering after sensory re-grounding. Additionally, in closed-loop hardware transfer, all policies succeed in a setting close to the simulated geometry, whereas the policy trained from a single fixed layout fails under a corridor-topology and obstacle-type shift, while the other three succeed.

\end{abstract}


%
%

\section{Introduction}
\begin{figure}[t]
    \centering
    \includegraphics[width=\linewidth]{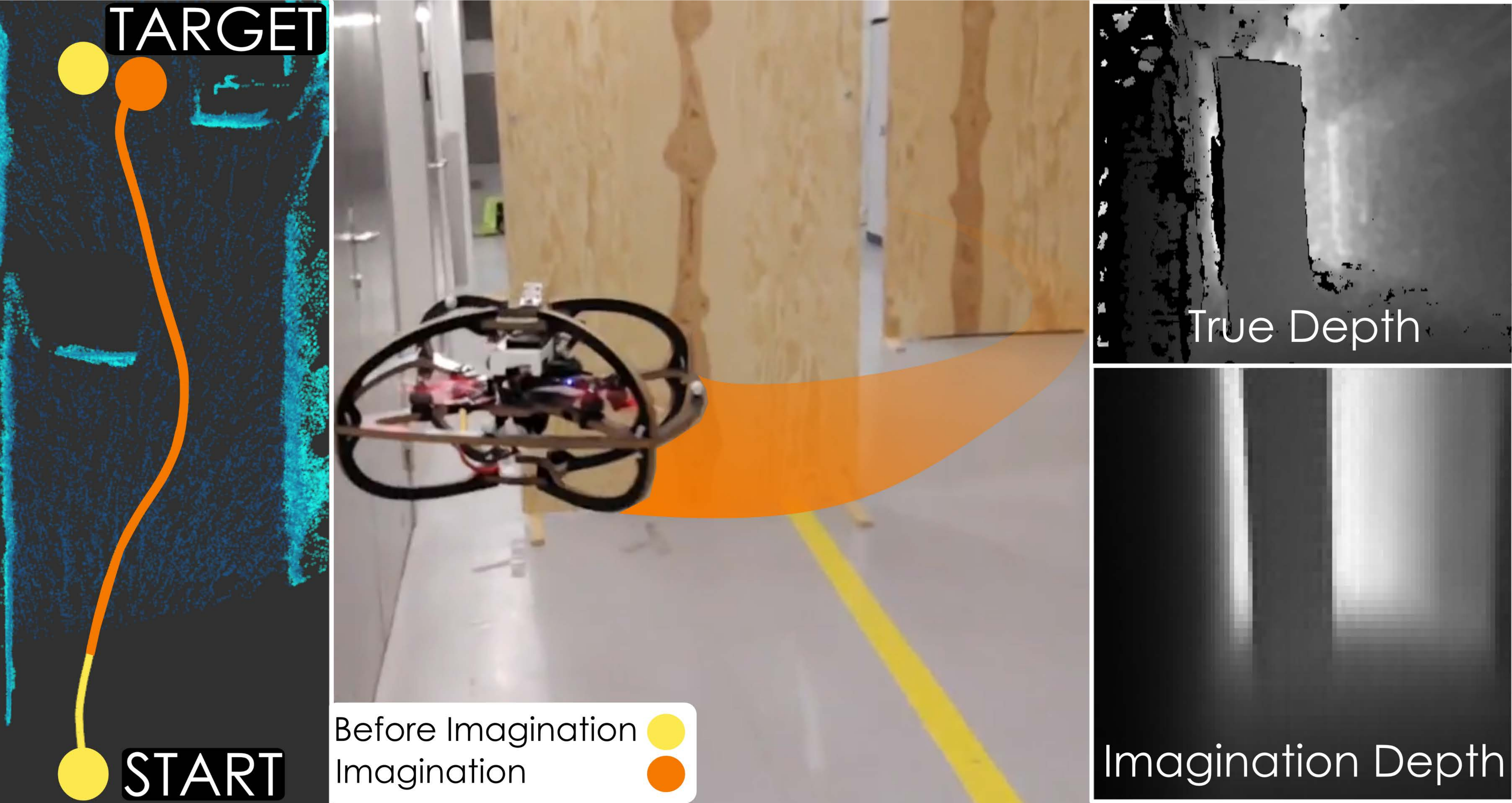}
    \vspace{-4ex}
    \caption{Once real sensing is cut, the quadrotor completes the flight to the target guided by the world model's imagined state and depth, avoiding obstacles. The environment is visualized as an accumulated point cloud.}
    \label{fig:title_figure}
     \vspace{-3ex}
\end{figure}

Advances in embodied AI emphasize predictive representation learning, where the agent learns an internal model of how the world evolves under its actions. According to this paradigm, \ac{mbrl} leverages a learned dynamics model, referred to as the world model, in order to synthesize future experience through latent imagination as opposed to exclusively relying on computationally expensive interaction with the environment~\cite{ha2018recurrent, janner2019trust}. Here, imagination refers to rolling the learned latent dynamics forward without new sensory observations, as shown in~\Cref{fig:title_figure}. 
By enabling agents to internally predict future outcomes, world models offer a pathway toward improved sample efficiency, long-horizon reasoning, and scalable decision-making in high-dimensional problems~\cite{hafner2019learning,hafner2020Dream,hafner2021mastering,hafner2025mastering, aljalbout2025accelerating, maes2026leworldmodel}. In fact, beyond policy optimization alone, world models demonstrate potential to serve as a more general computational primitive for embodied intelligence, supporting state estimation, trajectory prediction, and self-supervised representation learning across diverse robotic systems~\cite{schrittwieser2020mastering, wu2023daydreamer, hansen2024td, bar2025navigation, li2025robotic}.
Within this broad class, we focus specifically on DreamerV3-style reconstructive latent-dynamics models, whose reconstruction objective allows the learned predictions to be evaluated directly.

Despite progress, it remains unclear how DreamerV3-style predictive representations behave when the training and evaluation environment distributions differ. Understanding this behavior is important when such representations are used to train navigation policies.
Existing approaches typically evaluate within relatively fixed settings, implicitly assuming that training and deployment environments are highly similar~\cite{hafner2019learning,hafner2020Dream,hafner2021mastering,hafner2025mastering} or even require exact digital twins of the target environment~\cite{romero2025dream}. While these methods yield exciting results, their achieved control performance does not necessarily 
reveal how well the learned predictive representation transfers across the environments.
This limitation is particularly important for vision-based navigation of aerial robots deployed across highly variable real-world scenarios. In such tasks, predictive performance may degrade significantly due to environmental variability (structure, obstacle configuration, light conditions, texture). The problem is made even more demanding by the sudden viewpoint changes inherent to agile flight dynamics.


While recent works have explored applying world models to the autonomous navigation of aerial robots~\cite{romero2025dream,verraest2025skydreamer, geles2024demonstrating}, an important limitation of the current state-of-the-art remains: the lack of systematic analysis regarding environmental variability and generalization. 
In~\cite{romero2025dream}, the authors train separate world models for specific individual racing tracks and validate deployment in a hardware-in-the-loop setup, without investigating transfer across environments. The work in~\cite{verraest2025skydreamer} demonstrates sim-to-real transfer across two scenarios but mitigates the visual reality gap through specialized pre-processing of the onboard observations into binary gate masks before training the world model. 
These works motivate a more systematic analysis of how DreamerV3-style models respond to changes in the environment distribution, particularly when transferring vision-based navigation policies beyond the settings used for training.

To address this gap, we investigate depth-based quadrotor navigation with the DreamerV3 world model~\cite{hafner2025mastering}.
Our study uses four distinct environment-generation regimes that differ in layout support and obstacle-sampling strategy.
We examine reconstruction across these regimes during \ac{ssl} pretraining, followed by policy performance evaluation after \ac{rl} fine-tuning in simulation. 
Finally, we deploy the learned world models to a physical quadrotor to assess closed-loop performance under environmental shifts and evaluate open-loop flight maintained only through latent imagination during sensor-free intervals.
To our knowledge, this is the first demonstration of real-world quadrotor navigation in which onboard state and depth measurements are withheld, and flight is driven solely by imagined observations.

\subsection{Contributions}

Taken together, our contributions are: \emph{First}, we systematically evaluate DreamerV3-style reconstructive latent-dynamics models trained under four distinct environment-generation regimes, cross-evaluating reconstruction before and after \ac{rl} fine-tuning and separately measuring navigation-policy performance in simulation. \emph{Second}, we perform sim-to-real transfer of the resulting policies on a real quadrotor in both a simulation-like setting and under a shift in corridor topology and obstacle type, providing a complementary view of deployment behavior beyond simulation reconstruction or policy metrics alone. \emph{Third}, we provide the first demonstration of real-world quadrotor flight driven solely by imagined state and depth during sensor-free intervals, including repeated \SI{5}{\second} imagination windows with periodic sensory re-grounding. 
The method is open-sourced in \url{https://github.com/ntnu-arl/flying-on-imagination}.

\section{Problem Formulation}
\label{sec:problem_formulation}
This work investigates the generalization of DreamerV3-style predictive world models for autonomous collision-free 3D navigation in unknown environments using onboard depth perception.
Let us denote the inertial frame as $\mathcal{I}$, the body-fixed frame as $\mathcal{B}$. At time $t$, the robot state is: $\mathbf{s}_t = [\mathbf{p}_t, \mathbf{v}_t, \mathbf{R}_t, \boldsymbol{\omega}_t]$ ,where $\mathbf{p}_t =[p_{x,t}, p_{y,t}, p_{z,t}]^\top $, is the position defined in $ \mathcal{I}$, $\mathbf{v}_t = [v_x, v_y, v_z]^\top $ is the linear velocity expressed in $ \mathcal{I}$, $\mathbf{R}_t \in \mathrm{SO}(3)$ is the rotation matrix from $\mathcal{I}$ to $\mathcal{B}$, and $\boldsymbol{\omega}_t = [\omega_x, \omega_y, \omega_z]^\top $ is the angular velocity in $ \mathcal{B}$. The robot perceives its surroundings through a depth image $\mathbf{D}_t$, which together with $\mathbf{s}_t$ constitutes the full onboard observation available to the agent at each timestep.

Given a goal position $\mathbf{p}_{\text{goal}}$ in $\mathcal{I}$, the current state $\mathbf{s}_t$, and the depth image $\mathbf{D}_{t}$, the objective of this work is to learn a control policy $\pi$ that outputs velocity commands $\mathbf{a}_t = [a_{v_x}, a_{v_y}, a_{v_z}, a_{\omega_z}]^\top$ in $\mathcal{B}$, comprising a 3D linear velocity command and a yaw rate, such that the robot reaches $\mathbf{p}_{\text{goal}}$ while avoiding collisions with obstacles. To keep motion within the camera field of view, we constrain the action space: the lateral velocity is fixed to $a_{v_y} = 0$, and the longitudinal and vertical commands $a_{v_x}, a_{v_z}$ are projected onto the maximum inclination angle.

Beyond the closed-loop setting, where $\pi$ acts on real onboard observations $\{\mathbf{s}_t, \mathbf{D}_t\}$ at every timestep, we additionally study navigation \emph{in imagination}. After a short context window of real observations, all further sensory input is withheld, and $\pi$ instead acts on predicted observations $\{\hat{\mathbf{s}}_t, \hat{\mathbf{D}}_t\}$, propagated forward internally and without access to the real sensor stream.
\section{Method}
\label{sec:method}

\begin{figure*}[t]
    \centering
    \includegraphics[width=\linewidth]{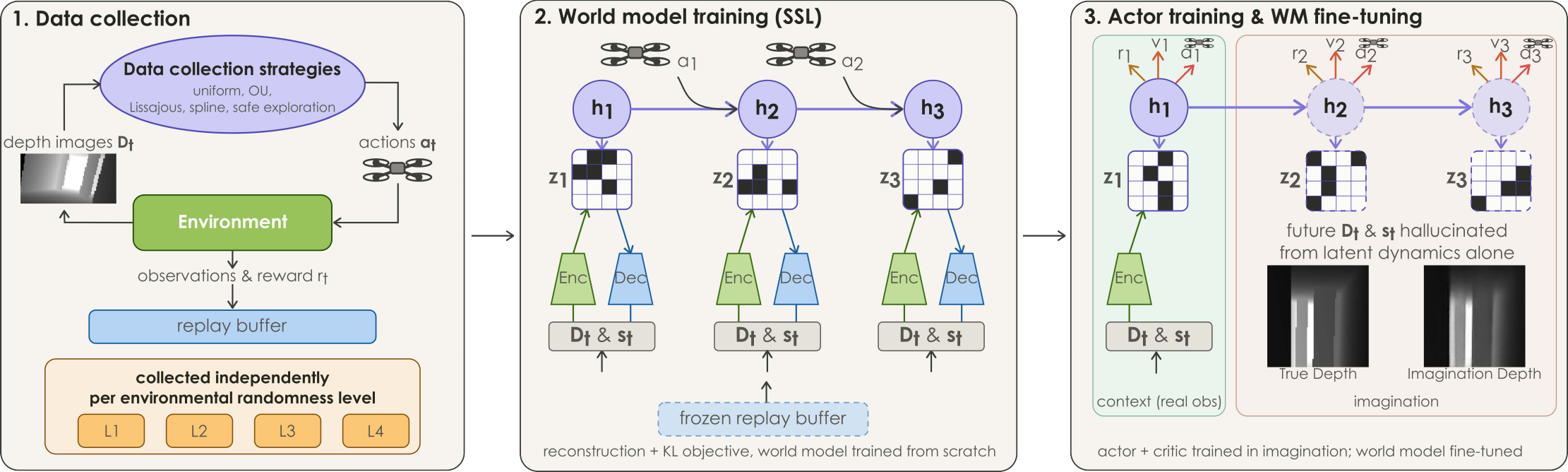}
    \vspace{-4ex}
    \caption{First, the data is collected per environmental regime. Next, the DreamerV3 world model is trained by self-supervised reconstruction. Finally, an actor and critic are trained in imagination with world-model fine-tuning: after a real-observation context (green), the model rolls out in pure imagination (orange), hallucinating future observations.
    }
    \label{fig:method}
    \vspace{-3ex}
\end{figure*}

The method comprises three stages: (\textbf{1}) data collection under different environmental-generation regimes (\Cref{sec:envs}), (\textbf{2}) \ac{ssl} pretraining of DreamerV3 world models with hyperparameter optimization (\Cref{sec:dreamer}), and (\textbf{3}) \ac{rl} fine-tuning of the best world model per environmental regime (\Cref{sec:actor}). The method overview is illustrated in~\Cref{fig:method}. 

\subsection{Data collection}
\label{sec:envs}
To investigate world model robustness to domain shift, we define four environment-generation regimes with different layout supports and obstacle-sampling strategies within a simulated indoor environment of $16 \times 7 \times \SI{7}{\m}$ containing five cuboid obstacles of size $0.1\times1.2\times\SI{7}{\m}$. All levels share the same environment geometry, but they differ in how obstacles spawn, as well as how goal positions are configured at episode initialization. Each level $\mathrm{L}i$ defines an environment configuration, and we denote the world model trained on data collected in $\mathrm{L}i$, together with its fine-tuned policy, as $\mathrm{WM}i$ throughout the paper. In the first setting, $\mathrm{L}1$, obstacles are in a fixed configuration with only the spawn and goal positions sampled uniformly at each episode. $\mathrm{L}2$ introduces structured diversity by selecting at each episode from five distinct fixed obstacle configurations, one of which coincides with the $\mathrm{L}1$ placement. $\mathrm{L}3$ introduces continuous spatial variation by drawing obstacle positions from a Sobol quasi-random sequence~\cite{sobol1967distribution}, providing low-discrepancy coverage of the environment space. It ensures that obstacles are distributed more uniformly across the environment than pure random sampling would achieve. Lastly, $\mathrm{L}4$ samples obstacles uniformly and independently at every episode. 

We choose four regimes to span qualitatively different environment-generation processes while keeping the full cross-evaluation computationally tractable.
In particular, $\mathrm{L}3$ introduces continuous obstacle variation while maintaining more even spatial coverage, whereas the independent uniform sampling of $\mathrm{L}4$ can produce clustered obstacle configurations. This distinction allows us to compare two continuously varying training distributions with different spatial-sampling properties.


Since no pre-trained policy exists that can be leveraged for training, we collect trajectories for \ac{ssl} using a set of exploration strategies. High-frequency strategies generate temporally diverse action sequences that cover the action space: uniform random sampling, \ac{ou} noise~\cite{uhlenbeck1930theory}, Lissajous curves, and a pseudo-spline smooth random walk. As a low-frequency strategy, we use a reactive depth-based controller we term \emph{safe exploration}. This controller steers the quadrotor by comparing mean depth across left, center, and right horizontal image sectors, applying corrective yaw commands when the forward sector approaches a proximity threshold. All these strategies yield trajectories spanning qualitatively distinct flight regimes, from highly stochastic motion to geometrically structured traversals. Finally, the data is collected independently for each level $\mathrm{L}i$. For clarity, we summarize: $\mathrm{WM}1$ is trained in a fixed-layout environment ($\mathrm{L}1$), $\mathrm{WM}2$ on episodes drawn from five fixed layouts ($\mathrm{L}2$), $\mathrm{WM}3$ on Sobol-distributed obstacle placements ($\mathrm{L}3$), and $\mathrm{WM}4$ on fully i.i.d.\ uniform placements ($\mathrm{L}4$). 

\subsection{World model training (\texorpdfstring{\ac{ssl}}{ssl})}
\label{sec:dreamer}
We build on DreamerV3~\cite{hafner2025mastering}, which learns a world model of the environment's dynamics and optimizes a policy entirely within latent imagination. At each step, the agent receives a depth image $\mathbf{D}_t$ and proprioceptive state $\mathbf{s}_t$, and outputs an action $\mathbf{a}_t$. The world model is a \ac{rssm} with a deterministic recurrent state $\mathbf{h}_t$ and a categorical stochastic state $\mathbf{z}_t$:
\begin{align*}
\text{recurrent model:} \quad & \mathbf{h}_t = f_\phi(\mathbf{h}_{t-1}, \mathbf{z}_{t-1}, \mathbf{a}_{t-1}), \\
\text{encoder:} \quad        & \mathbf{z}_t \sim q_\phi(\mathbf{z}_t \mid \mathbf{h}_t, \mathbf{D}_t, \mathbf{s}_t), \\
\text{dynamics predictor:} \quad & \mathbf{\hat z}_t \sim p_\phi(\mathbf{\hat z}_t \mid \mathbf{h}_t), \\
\text{decoder:} \quad        & (\mathbf{\hat D}_t, \mathbf{\hat s}_t) = g_\phi(\mathbf{h}_t, \mathbf{z}_t), \\
\text{reward predictor:} \quad & \hat r_t \sim p_\phi(\hat r_t \mid \mathbf{h}_t, \mathbf{z}_t),
\end{align*}
where $\phi$ denotes the world model's parameters. The latent state \{$\mathbf{h}_t$, $\mathbf{z}_t$\} encodes both memory and uncertainty, and the decoder grounds it in perceptual content, enabling imagined observations to be compared directly against real state and depth readings at deployment. The world model is trained to minimize a joint objective:
\begin{equation}
    \mathcal{L}_\phi = \mathcal{L}_{\text{rec}} + \mathcal{L}_{\text{rew}} + \beta\, \mathcal{L}_{\text{KL}},
\end{equation} 
where $\mathcal{L}_\text{rec}$ reconstructs image $\mathbf{D}_t$ and state $\mathbf{s}_t$. $\mathcal{L}_\text{rew}$ predicts the reward, while $\mathcal{L}_\text{KL}$ regularizes the stochastic state by aligning the encoder $q_\phi$ and dynamics predictor $p_\phi$. The hyperparameter $\beta$ balances reconstruction fidelity against KL regularization. This procedure is illustrated in the ``World model training'' stage of~\Cref{fig:method}.
For each training regime, a dedicated world model $\mathrm{WM}i$ is trained on the collected trajectories via \ac{ssl} by minimizing the loss $\mathcal{L}_\phi$.


\subsection{Actor training \& WM fine-tuning}
\label{sec:actor}
For each training regime $\mathrm{Lj}$, the corresponding world model is fine-tuned using \ac{rl}
entirely within latent imagination, as shown in the third stage of~\Cref{fig:method}. The actor policy $\pi_\theta$ is optimized to maximize the expected discounted cumulative reward:
\begin{equation*}
    J(\pi_\theta) = \mathbb{E}_{\pi_\theta}\left[\sum_{t=0}^{T} \gamma^t r_t\right]
\end{equation*}
where $\theta$ denotes the policy parameters, $T$ is the episode length, $r_t$ denotes the reward at time $t$, and $\gamma \in (0,1)$ is the discount factor. The reward function combines a proximity term with a time-efficiency bonus:
\begin{equation*}
    r_t = 2\,e^{-\|d_t\|^2} + 2\,e^{-0.008\,\|d_t\|^2} + \mathbb{I}(\|d_t\| < 0.5) \cdot (T - t)
\end{equation*}
where $d_t$ is the Euclidean distance to the goal at time $t$, and $\mathbb{I}(\cdot)$ is an indicator function evaluating to $1$ when the quadrotor is within goal proximity ($\|d_t\| < \qty{0.5}{\m}$), and $0$ otherwise. The first two exponential terms provide a dense guidance signal at both short and long ranges, respectively. The indicator term yields a larger bonus of $T - t$ when the goal is reached at an earlier step $t$, directly rewarding navigation efficiency. Explicit crash and success bonuses are set to zero, so the agent is shaped entirely through proximity and time efficiency.

The actor unrolls trajectories through the \ac{rssm} and the critic estimates the value function $V_\psi(\mathbf{h}_t, \mathbf{z}_t)$ along the imagined rollouts. The actor is trained by backpropagating gradients directly through the imagined trajectory using \ac{ema}-normalized targets:
\begin{equation*}
\mathcal{L}_\text{actor} = -\mathbb{E}_{\tau \sim \pi_\theta}\left[\sum_{t=0}^{H-1} w_t \cdot \frac{V^\lambda_t - \mu}{\sigma}\right]    
\end{equation*}
where $H$ is the imagination horizon, $V^\lambda_t$ is the $\lambda$-return, $w_t$ are importance weights, and $\mu$ and $\sigma$ are running \ac{ema} statistics of the return distribution. To introduce control-level uncertainty during \ac{rl} fine-tuning, the parameters of the flying vehicle controller~\cite{lee2010geometric}, used as the low-level dynamics model, are randomized at each episode. This isolates the effect of dynamics mismatch on policy robustness independently of visual or obstacle-placement variability~\cite{tobin2017domain, peng2018sim}. 

\subsection{Implementation Details}
All training is conducted in AerialGym~\cite{kulkarni2025aerial}. Depth observations have resolution $64\times96$ and are sampled together with the robot state at \SI{10}{\Hz}. Thus, the longest tested sequence, $L_{\mathrm{batch}}=64$, spans \SI{6.4}{\s}.
For cross-environment reconstruction, each world model $\mathrm{WM}i$, trained on regime $Li$, is evaluated on held-out trajectories from every regime $Lj$.
Each evaluation sequence is conditioned on real observations for \SI{1}{\s}, after which the model is rolled out open-loop from its latent dynamics and compared with the corresponding ground-truth observations.
\ac{rl} policies are trained for $1$M actor interactions with the environments and evaluated over $200$ episodes per model–environment pair, including an \ac{ood} environment with $10$ randomly placed cuboids. \ac{ppo} is trained independently on each $Lj$ using the identical reward function and evaluated using the same navigation criteria.


\begin{figure*}[t]
        \centering
        \includegraphics[width=\linewidth]{images/newPlot.pdf}
        \vspace{-4ex}
        \caption{Hyperparameter sweep results per environment ($\mathrm{L}1$-$\mathrm{L}4$): each cell shows all runs per environment (grey) with the best configuration highlighted. Stages S1–S4 sweep \ac{rssm} size, batch length, discrete latent size, and seed, respectively, in sequence. 
        }
        \label{fig:multi_env_curves}
        \vspace{-3ex} 
\end{figure*}
\section{Results}
\label{sec:results}
Our evaluation follows three stages. We first examine self-supervised pretraining, analyzing hyperparameter sensitivity and cross-environment reconstruction quality in~\Cref{sec:ssl_results}. Then, we evaluate \ac{rl} fine-tuning in terms of policy performance and imagination fidelity in~\Cref{sec:rl_results}. Finally, we present sim-to-real deployment of all models on a real quadrotor across both simulation-like setups and novel environment shifts, testing closed-loop execution and open-loop imagination in~\Cref{sec:real_world}. 

\subsection{Self-Supervised Pretraining Analysis}
\label{sec:ssl_results}
In the first ablation study, we evaluate sensitivity to four important DreamerV3 hyperparameters: \ac{rssm} deterministic-state size $d_{\rm det}$, network hidden size $d_{\rm hid}$, training-sequence length $L_{\rm batch}$, and discrete latent size $d_{\rm disc}$. The first sequential sweep $\mathrm{S}1$ uses $d_{det} \in \{256, 512, 1024\}$ and $d_{hid} \in \{256, 512, 1024\}$ of the \ac{rssm}, fixing the remaining parameters to their defaults. We carry forward the configuration that achieves the lowest evaluation loss to the next phase $\mathrm{S}2$.  There, we sweep the sequence length $L_{batch} \in \{16, 32, 64\}$. Next, stage $\mathrm{S}3$ sweeps the $d_{disc} \in \{0, 32, 64\}$, where $0$ means normal distribution, hence no categorical. Finally, phase $\mathrm{S}4$ re-runs the best overall configuration with five different random seeds to confirm that the results are not seed-dependent. At each stage, candidates are selected by reconstruction loss on held-out trajectories from the corresponding training regime. The selected model is subsequently cross-evaluated on all four regimes in both context and imagination.

\begin{figure}[t]
        \centering
        \includegraphics[width=\columnwidth]{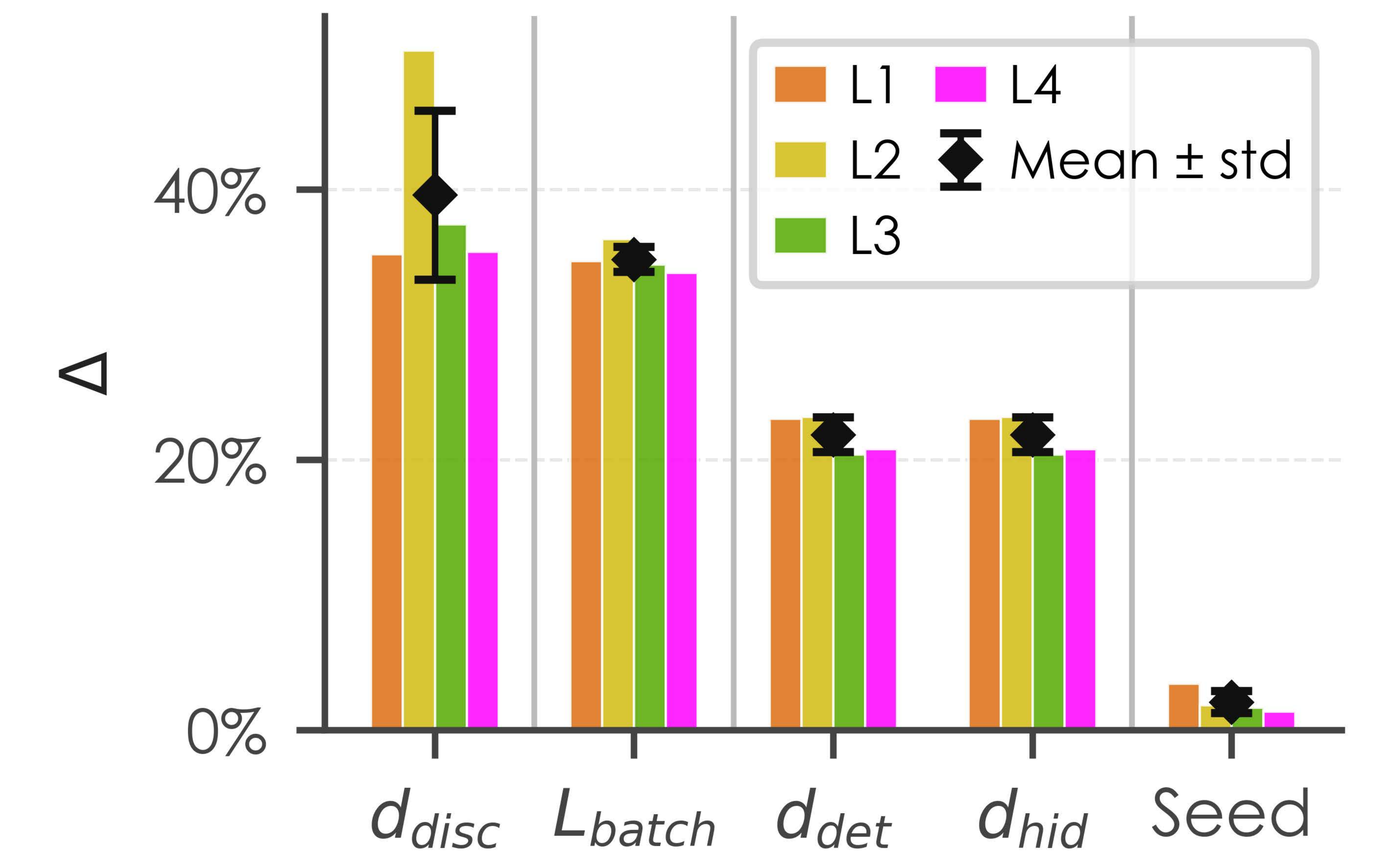}
        \vspace{-4ex}
        \caption{Hyperparameter sensitivity across environments. The relative loss gap $\Delta$ between the worst and best value of each hyperparameter, per environment ($\mathrm{L}1$-$\mathrm{L}4$).}
        \label{fig:multi_env_importance}
        \vspace{-3ex}
\end{figure}

\begin{figure*}[t]
    \centering
    \includegraphics[width=\linewidth]{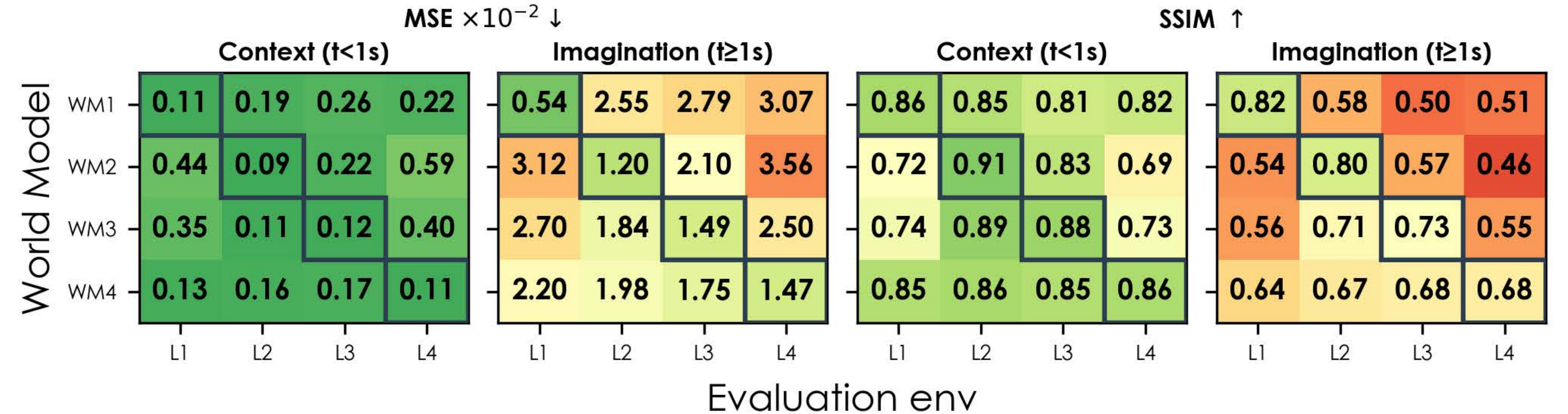}
    \vspace{-4ex}
    \caption{Cross-environment validation heatmap (MSE$\downarrow$ / SSIM$\uparrow$) reports reconstruction quality for world model $\mathrm{WM}i$ evaluated on environment $\mathrm{L}j$, split into context and imagination. Columns fix the evaluation regime and are used for cross-model comparisons.
    }
    \label{fig:heatmaps}
     \vspace{-3ex}
\end{figure*}

\begin{figure}[t]
    \centering
    \includegraphics[width=\linewidth]{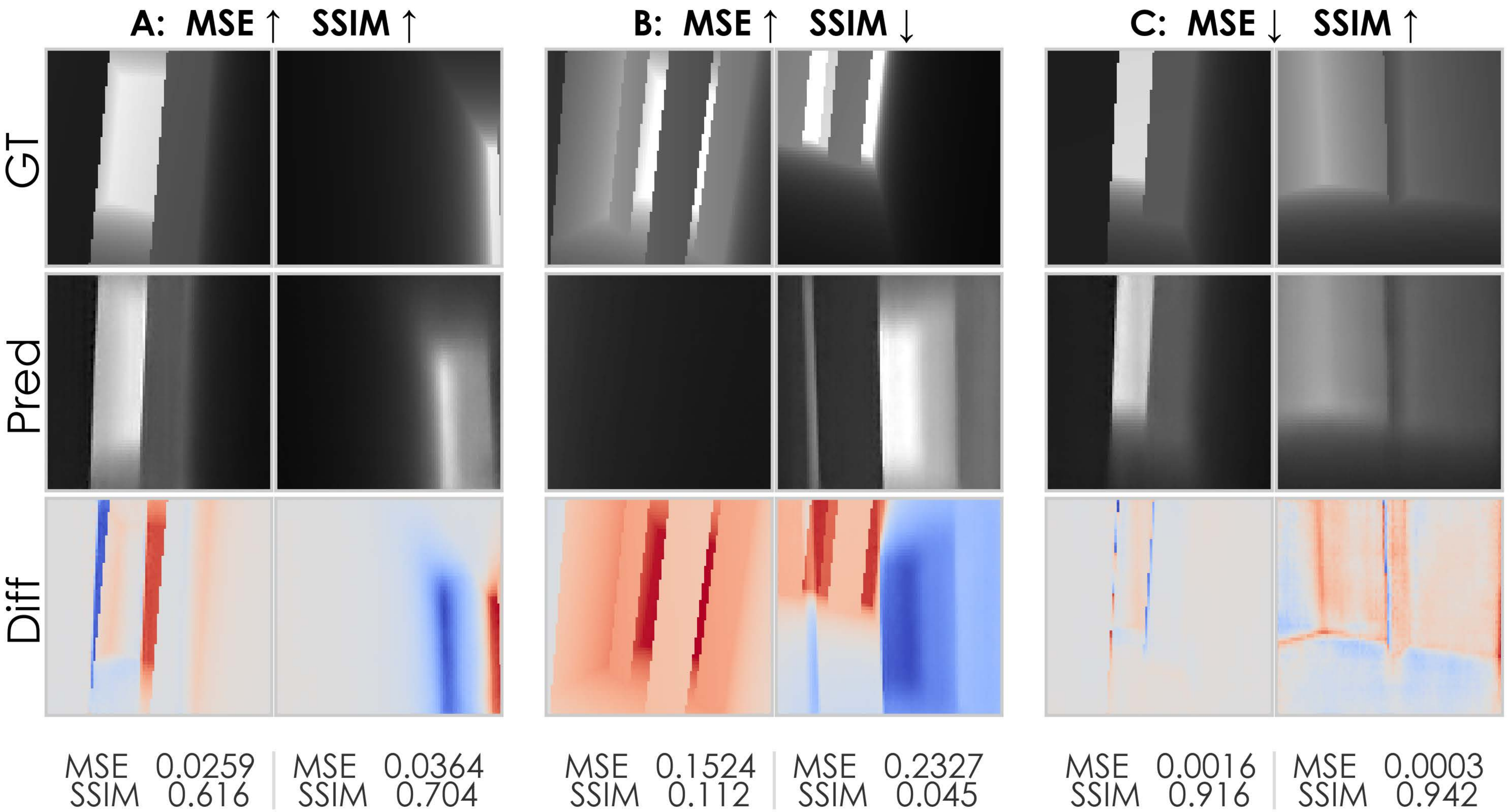}
     \vspace{-3ex}
    \caption{Qualitative illustration of \ac{mse} and \ac{ssim} as complementary reconstruction metrics. Each panel shows a GT/Pred/Diff triplet for a representative case: (A) high \ac{mse}  yet high \ac{ssim}, where pixel-level errors are present but structural patterns are preserved; (B) high \ac{mse}  and low \ac{ssim}, indicating both pixel-level and structural failure; (C) low MSE and high \ac{ssim}, corresponding to accurate reconstruction of both intensity and structure.}
    \label{fig:metrics_app}
    \vspace{-3ex}
\end{figure}

The results are shown in~\Cref{fig:multi_env_curves}. In every cell, the best configuration (colored) converges to a clearly lower loss than the remaining candidates (grey), and this margin holds across all four environments. The best configurations uniformly favor large \ac{rssm} capacity ($d_\text{det} = d_\text{hid} = 1024$) and long batch sequences ($L_\text{batch} = 64$), with $d_\text{disc} = 64$ for $\mathrm{L}1$/$\mathrm{L}2$ and $d_\text{disc} = 32$ for $\mathrm{L}3$/$\mathrm{L}4$. This suggests that the preferred latent bottleneck may depend on the training distribution, although the sequential sweep does not isolate the cause of this difference. The configurations are consistent across the environmental regimes: world model quality is governed far more by architectural and training choices than by the environment it is trained in, so the same configuration can be reused across environments without per-environment re-tuning.

In~\Cref{fig:multi_env_importance}, we report the relative impact of each hyperparameter on evaluation loss, $\Delta = \frac{\ell_\text{worst}-\ell_\text{best}}{\ell_\text{worst}}$, across all four environments. Along the sequential sweep, $d_\text{disc}$ and batch length $L_\text{batch}$ show the largest observed relative loss variations, but in different ways: $L_\text{batch}$ contributes a stable $\Delta\approx35\%$ across regimes, whereas $d_\text{disc}$ shows larger and more variable changes, ranging from $\approx35\%$ to $\approx50\%$ across the four regimes. The \ac{rssm} sizes $d_\text{det}$ and $d_\text{hid}$ have a moderate effect ($\Delta\approx20\%$), while seed variance is negligible ($\Delta<5\%$) relative to the changes observed along the hyperparameter sweep, confirming that the differences reflect architectural sensitivity rather than training noise. 

\begin{table}[t]
\centering
\caption{RL Cross-Environment Policy Evaluation (200 Episodes). Diagonal (In-Distribution) Entries are \underline{Underlined}.}
\vspace{-3ex}
\label{tab:rl_policy}
\footnotesize
\setlength{\tabcolsep}{4.5pt}
\renewcommand{\arraystretch}{1.1}
\begin{tabular}{ll rrrrr}
\toprule
Metric (\%) & Model & $\mathrm{L}1$ & $\mathrm{L}2$ & $\mathrm{L}3$ & $\mathrm{L}4$ & OOD \\
\midrule
& $\mathrm{WM}1$ & \underline{99.5} & 89.5 & 72.5 & 75.0 & 54.5 \\
Success Rate ($\uparrow$) & $\mathrm{WM}2$ & 88.0 & \underline{88.5} & 87.5 & 90.0 & 55.5 \\
& $\mathrm{WM}3$ & 97.0 & 96.5 & \underline{92.5} & 89.5 & 72.0 \\
& $\mathrm{WM}4$ & 91.5 & 95.0 & 95.0 & \underline{88.0} & 67.0 \\
\cmidrule{2-7}
& $\mathrm{WM}1$ & \underline{0.5} & 10.5 & 27.0 & 25.0 & 44.5 \\
Crash Rate ($\downarrow$) & $\mathrm{WM}2$ & 12.0 & \underline{11.5} & 12.5 & 10.0 & 44.5 \\
& $\mathrm{WM}3$ & 3.0 & 3.5 & \underline{7.5} & 10.5 & 27.0 \\
& $\mathrm{WM}4$ & 8.5 & 5.0 & 5.0 & \underline{11.5} & 32.5 \\
\cmidrule{2-7}
& $\mathrm{WM}1$ & \underline{0.0} & 0.0 & 0.5 & 0.0 & 1.0 \\
Timeout Rate ($\downarrow$) & $\mathrm{WM}2$ & 0.0 & \underline{0.0} & 0.0 & 0.0 & 0.0 \\
& $\mathrm{WM}3$ & 0.0 & 0.0 & \underline{0.0} & 0.0 & 1.0 \\
& $\mathrm{WM}4$ & 0.0 & 0.0 & 0.0 & \underline{0.5} & 0.5 \\
\bottomrule
\end{tabular}
\end{table}

\begin{table}[t]
\centering
\caption{Success Rate (\%) at Matched Training Budget: World Model (1M) vs.\ PPO Baselines across Environments.}
\vspace{-3ex}
\label{tab:ppo}
\footnotesize
\setlength{\tabcolsep}{5pt}
\renewcommand{\arraystretch}{1.15}
\begin{tabular}{l ccc r}
\toprule
& \textbf{World Model} & \multicolumn{2}{c}{\textbf{PPO Baseline}} & \multirow{2}{*}{\shortstack{\textbf{PPO Steps}\\\textbf{to Match WM}}} \\
\cmidrule(lr){2-2} \cmidrule(lr){3-4}
\textbf{Level} & In-Dist. (1M) & 1M Steps & 15M Steps & \\
\midrule
$\mathrm{L}1$ & 99.5 & 0.0 & 99.6 & 4.0M \\
$\mathrm{L}2$ & 88.5 & 0.0 & 98.9 & 7.5M \\
$\mathrm{L}3$ & 92.5 & 0.0 & 95.4 & 5.9M \\
$\mathrm{L}4$ & 88.0 & 0.0 & 91.7 & 1.9M \\
\bottomrule
\end{tabular}
\vspace{-3ex}
\end{table}

\Cref{fig:heatmaps} presents \ac{mse} and \ac{ssim} for each world model evaluated across all environments, split into the context phase ($t < 1\,\text{s}$, conditioned on real observations) and the imagination phase ($t \geq 1\,\text{s}$, purely hallucinated). We report both metrics throughout, as they capture complementary aspects of reconstruction quality: \ac{mse} measures pixel-wise depth error, whereas \ac{ssim} measures structural similarity. The two may disagree when walls and obstacle edges are placed correctly, but absolute depth values are offset. This behavior can be clearly seen in~\Cref{fig:metrics_app}. During context, all models achieve low \ac{mse} and high \ac{ssim} on the diagonal, confirming that each world model fits its training distribution well. Performance degrades off-diagonal, particularly for $\mathrm{WM}1$ and $\mathrm{WM}2$ when evaluated on the continuously sampled $\mathrm{L}3$/$\mathrm{L}4$ regimes, reflecting the limited diversity of their training data. 
Because diagonal entries change both the trained model and the evaluation regime, they do not provide a controlled comparison of model generalization. We therefore compare models primarily within columns, where the evaluation regime is fixed. For example, under $\mathrm{L}4$ evaluation, imagination MSE decreases from $3.07$/$3.56$ for $\mathrm{WM}1$/$\mathrm{WM}2$ to $2.50$/$1.47$ for $\mathrm{WM}3$/$\mathrm{WM}4$, indicating better reconstruction of this regime by models trained with continuous obstacle sampling.

\subsection{Actor Training \& World Model finetuning}
\label{sec:rl_results}


\begin{figure*}[t]
    \centering
    \includegraphics[width=\linewidth]{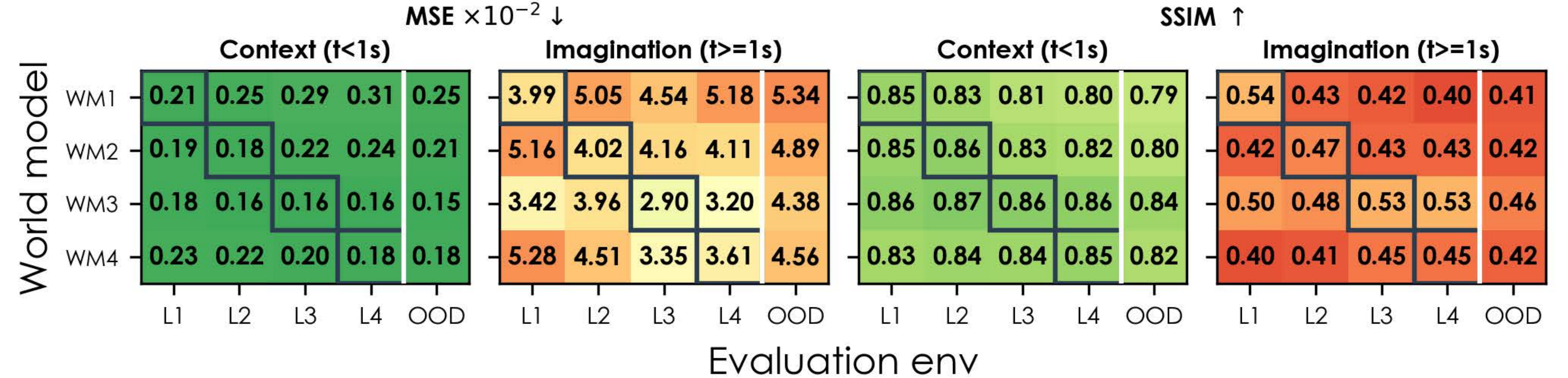}
    \vspace{-4ex}
    \caption{\ac{rl} imagination cross-environment metrics (MSE$\downarrow$ / SSIM$\uparrow$) reports reconstruction for world model $\mathrm{WM}i$ evaluated on environment $\mathrm{L}j$, split into context and imagination. 
    }
    \label{fig:metrics}
    \vspace{-3ex}
\end{figure*}
\begin{figure}[t]
    \centering
    \includegraphics[width=\linewidth]{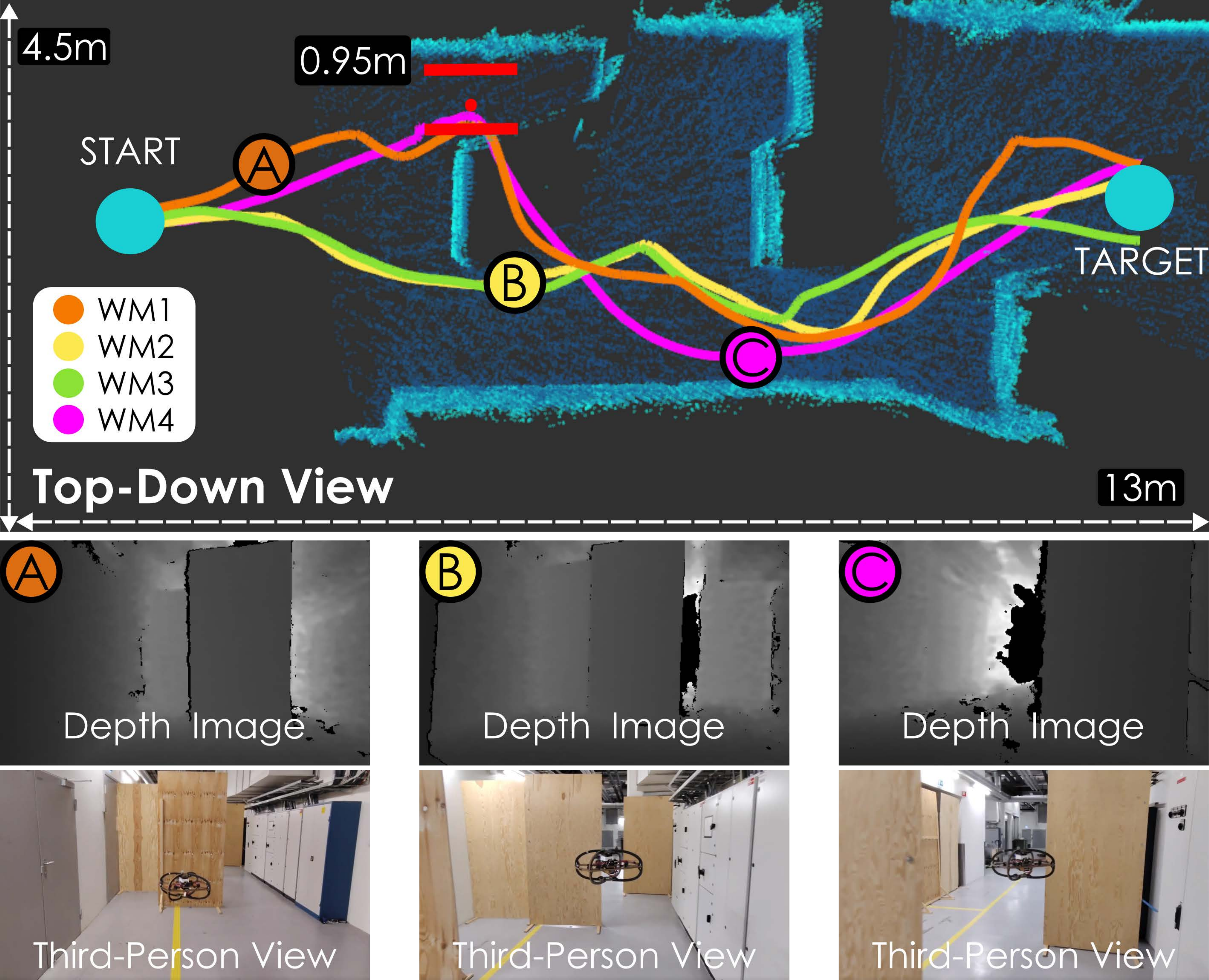}
    \vspace{-4ex}
    \caption{Closed-loop real-world deployment with 5 cuboid obstacles. Top-down trajectories of all trained world models. All trained world models successfully reach the target. Insets show onboard depth and third-person views. The red marker indicates the narrowest gap in this setting.}
    \label{fig:appendix_5cuboid}
    \vspace{-3ex}
\end{figure}

In~\Cref{tab:rl_policy}, we report the success, crash, and timeout rates for each world model policy, evaluated across all training environments and one \ac{ood} layout (a randomly sampled 10-cuboid environment). A success is defined as reaching the target proximity within the episode length, a crash as any contact with the environment, and a timeout otherwise. Within the training distribution, $\mathrm{WM}1$ achieves the highest success rate on its native environment ($99.5\%$). However, its performance degrades sharply when evaluated on the broader $\mathrm{L}2-\mathrm{L}4$ regimes and the \ac{ood}, with the crash rate rising from $0.5\%$ to $44.5\%$ in the \ac{ood} case. This reflects the limited diversity of the $\mathrm{L}1$ training data: the policy overfits to the fixed obstacle structure rather than learning a general avoidance strategy. In the evaluated run, $\mathrm{WM}3$, trained on Sobol-sampled obstacle placements, achieves success rates of at least 89.5\% across the four training regimes and $72.0\%$ on the more challenging \ac{ood} layout. Notably, $\mathrm{WM}4$ does not outperform $\mathrm{WM}3$ despite its more diverse training environment, suggesting that maximal stochasticity introduces variability that hampers policy convergence during \ac{rl} fine-tuning. Across all models, timeout rates are negligible, indicating that failures are exclusively collisions rather than navigation stagnation. 

\begin{figure}[t]
  \centering
  \includegraphics[width=\linewidth]{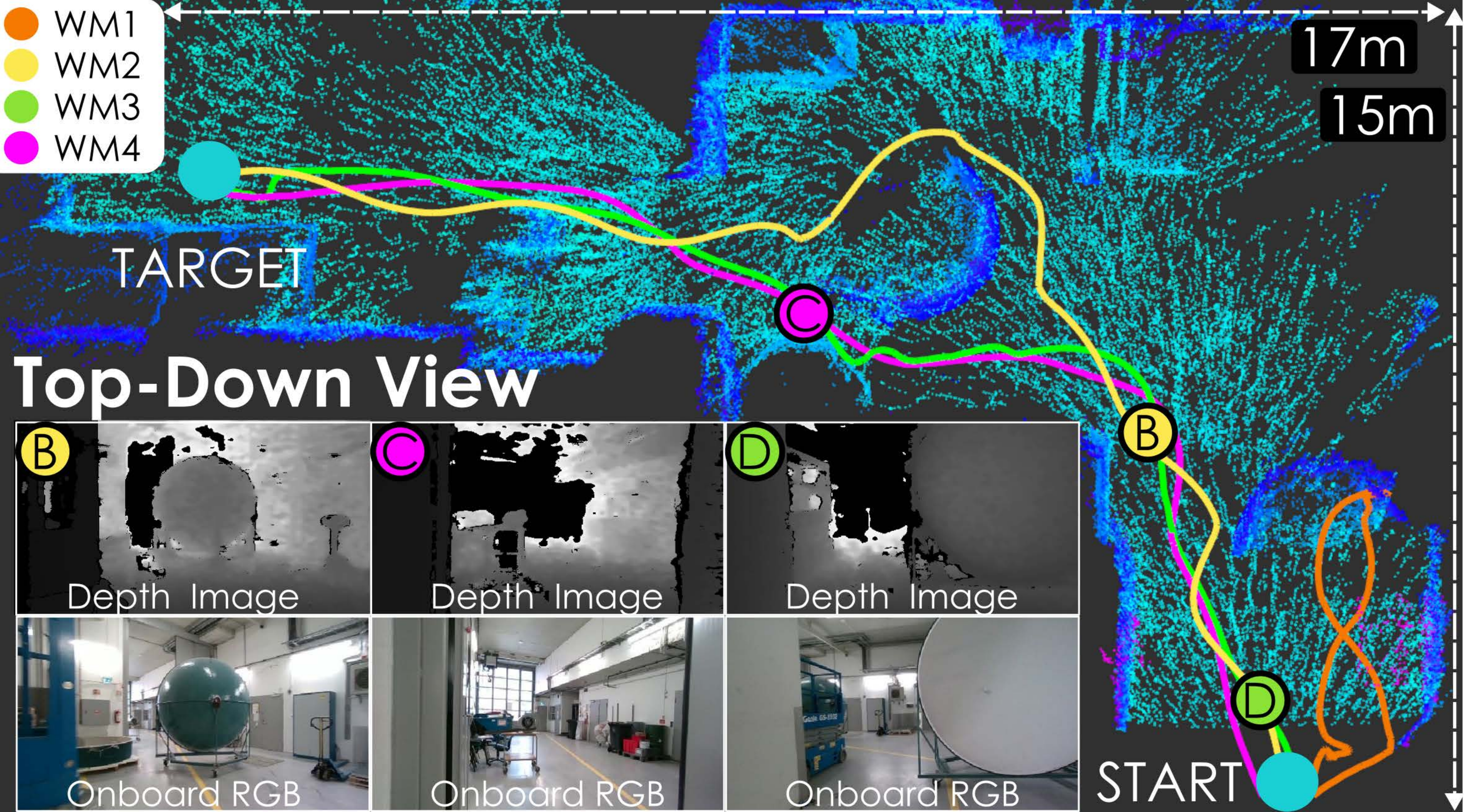}
  \vspace{-4ex}
  \caption{Closed-loop L-corridor flight with industrial (non-panel) obstacles. Insets \textbf{B}–\textbf{D} present onboard depth/RGB showcasing obstacle diversity.}
  \label{fig:closed-loop}
  \vspace{-3ex}
\end{figure}

\paragraph{Comparison with model-free baseline}
For comparison against the baseline approach, we train a \ac{ppo} policy on each $\mathrm{L}j$. The results are shown in~\Cref{tab:ppo}. Crucially, at 1M actor-interaction steps, \ac{ppo} has not yet learned to navigate and needs approximately 1.9 to 7.5M iterations (depending on the environment  $\mathrm{L}j$)  to match $\mathrm{WM}$'s success rate. This reflects \ac{mbrl}'s sample-efficiency advantage over the baseline~\ac{ppo} approach.


\paragraph{Post-RL latent reconstruction analysis}
\Cref{fig:metrics} reports reconstruction after \ac{rl} fine-tuning; it is not the \ac{rl}-performance metric. Because the checkpoint is selected by policy success rate rather than reconstruction loss, post-\ac{rl} \ac{mse}/\ac{ssim} should be read only as diagnostics of how the fine-tuned world model represents observations. Relative to the \ac{ssl} checkpoints, reconstruction degrades mainly over imagined rollouts while context-phase reconstruction remains more stable.
Notably, $\mathrm{WM}1$ shows the sharpest drop in \ac{ssim} under \ac{ood} evaluation, reinforcing that low training randomness limits generalization across both policy and world model dimensions.

\begin{figure*}[t]
  \centering
  \includegraphics[width=\linewidth]{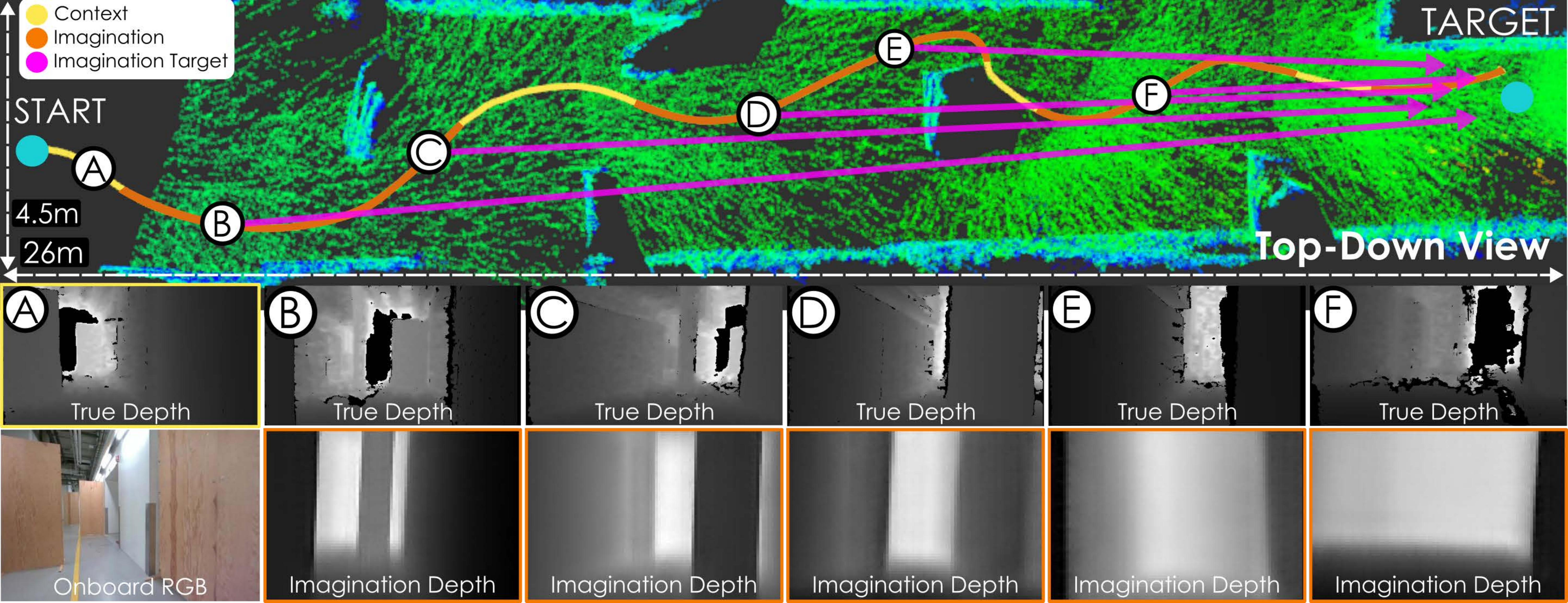}
  \vspace{-4ex}
  \caption{Real-world alternating context/imagination flight along a \SI{26}{\m}, four-panel corridor (yellow: context; orange: imagination). \textbf{B}–\textbf{F}: real vs.\ imagined depth, degrading with window length and recovering once context resumes.}
  \label{fig:imagination4}
  \vspace{-3ex}
\end{figure*}

\begin{figure}[t]
    \centering
        \includegraphics[width=\linewidth]{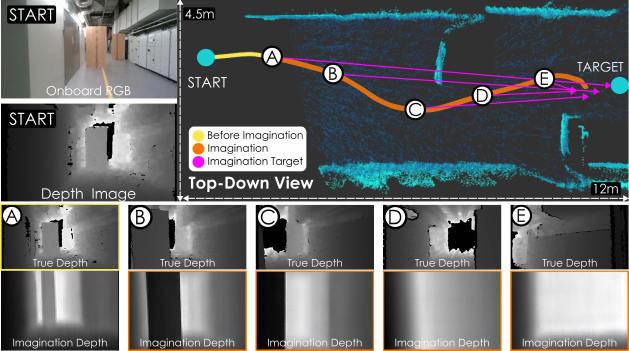}
        \vspace{-4ex}
        \caption{Open-loop imagination in the real world. After \SI{2.5}{\s} of context (yellow), all sensors are cut off (marker~\textbf{A}), and the robot flies in imagination (orange). The target position with magenta arrows and image insets contrasts the true sensor state and depth reading with the imagined rollout.
        }
        \label{fig:imagination1}
    \vspace{-3ex}
\end{figure}

Within simulation, \ac{ssl} reconstruction quality and \ac{rl} policy performance are related but not equivalent: $\mathrm{WM}4$ has the strongest \ac{ssl} cross-environment reconstruction, whereas $\mathrm{WM}3$ attains the strongest policy success rates.

\subsection{Real-World Deployment}
\label{sec:real_world}
We perform the real-world deployment on a custom-built quadrotor equipped with a VectorNav VN-100 \ac{imu} and a Texas Instruments IWR6843AOP radar sensor for state estimation~\cite{nissov2024degradation}, and an Intel RealSense D455 for depth perception. World models run onboard an NVIDIA Jetson Orin NX \SI{16}{\giga\byte}, while a PX4 autopilot handles low-level control. We perform the sim-to-real transfer without any fine-tuning for the deployed platform.



\paragraph{Closed-loop evaluation in simulation-like setting} First, we examine the closed-loop performance of the world models' policies. We populate the corridor with five panel-like obstacles, as shown in~\Cref{fig:appendix_5cuboid}. This matches the obstacle count used in simulation. To assess policy performance in this environment, we ask each model to perform a navigation to the goal location, which is \SI{12}{\meter} away from the start. We keep the starting and target positions fixed. For each policy, we repeat the task three times, and all four policies achieve a $100\%$ success rate while traversing even a gap as small as \SI{0.95}{\meter}, which is less than two robot sizes. This demonstrates that each learned world model's policy can transfer to real quadrotor navigation zero-shot when the deployment geometry remains close to the training setting.

\paragraph{Closed-loop transfer under topological and obstacle shift} Next, we deploy world model policies for three trials in an L-shaped corridor. This scenario represents a different environmental shape compared to the training scenario. It consists of industrial, non-panel obstacles, as shown in~\Cref{fig:closed-loop}. The models trained on more diverse environmental representations, namely $\mathrm{WM}2-\mathrm{WM}4$, succeed with a $100\%$ success rate, while $\mathrm{WM}1$ fails in every trial. $\mathrm{WM}1$ experiences only a single fixed layout ($\mathrm{L}1$) during training, whereas other models encounter multiple or continuously varying layouts. This result shows that training DreamerV3 models for vision-based navigation with environmental diversity allows for successful sim-to-real transfer in novel environments featuring geometrically comparable obstacles.


\paragraph{Open-loop continuous imagination} The second part of the real-world evaluation focuses on flying with a quadrotor on imagined states and depth observations. First, we test in an environment similar to the one used in training, composed of two panels, as presented in~\Cref{fig:imagination1}. The agent is again tasked to reach the target, which is \SI{12}{\meter} away from the starting location. The model begins the navigation towards the goal and receives \SI{2.5}{\s} of sensory context. The end of this window is indicated by marker A in~\Cref{fig:imagination1}. From this moment, the policy receives no further sensory input and rolls out future depth and state entirely from its latent dynamics. The indicative markers (A$\to$E) show the imagined depth along the image captured with the onboard depth camera as well as the imagined target location with magenta arrows. Early predictions closely match the real depth stream, correctly hallucinating corridor walls and obstacle edges, while fidelity gradually degrades over longer horizons. Despite this, the flying platform finishes the navigation near the true target location while maintaining collision-free motion. 

\paragraph{Alternating imagination with sensory re-grounding} The second imagination test aims to demonstrate the other ability of the world model policies, namely alternating between imagination and true sensor input. Now, the agent is tasked with navigating collision-free along a \SI{26}{\meter}-long corridor filled with five obstacles, as seen in the top-down view of~\Cref{fig:imagination4}. Here, we provide \SI{2.5}{\second} context windows followed by \SI{5}{\second} pure imagination windows. After each sensing cut, both state and depth are hallucinated from the latent state and used to compute the control command, which allows the robot to successfully navigate collision-free up to the true goal location. The robot traverses \SI{24.7}{\meter} in \SI{27.9}{\second}, spending $66\%$ of the time and $76\%$ of the distance in complete imagination. The post-processed reconstruction error grows with the open-loop horizon: \ac{mse} rises from $0.030$ in the first \SI{0.5}{\second} to $0.048$ beyond \SI{5}{\second}, and prediction quality recovers after sensory context is restored.


\section{Limitations}
\label{sec:limitation}


Our study is explicitly focused on DreamerV3-style reconstructive latent-dynamics world models, and thus the observed trends may not necessarily extend to other world-model architectures. The considered environmental variability is primarily geometric and control-level, and accordingly other sources of domain shift, such as sensing conditions, are not systematically varied.
This controlled scope allows us to isolate how changes in the training-environment distribution affect reconstruction, policy performance, and sim-to-real transfer.
Moreover, the real-world evaluation covers a limited set of indoor scenarios, so observations should be interpreted within the tested distribution shifts. Nevertheless, the successful deployment of policies trained under multiple environment-generation regimes, together with sensor-free flight driven by imagined state and depth, shows that the learned latent dynamics can remain useful beyond the exact simulation conditions. Finally, the imagination experiments show that prediction errors accumulate with open-loop horizon, motivating periodic sensory re-grounding for longer deployments.

\section{Conclusion}
\label{sec:conclusion}
We presented a systematic study of DreamerV3-style world models across different training-environment regimes for depth-based quadrotor navigation. Cross-environment \ac{ssl} evaluation shows how training diversity affects reconstruction quality, while \ac{rl} evaluation characterizes policy performance. On a real flying platform, all policies transfer successfully to a five-panel setting close to the simulation configuration, whereas $\mathrm{WM}1$, trained on a single fixed layout, fails in the L-corridor with novel obstacle types while $\mathrm{WM}2-\mathrm{WM}4$ succeed, providing evidence that training diversity improves robustness under the tested distribution shift. 
Finally, real-world imagination experiments demonstrate uninterrupted sensor-free flight, with all $\mathrm{WM}1$-$\mathrm{WM}4$ completing the open-loop navigation task and reaching the goal proximity without collision, as well as repeated \SI{5}{\s} imagined rollouts over a longer corridor, while showing prediction error accumulation with rollout horizon and recovery after sensory re-grounding.

\bibliographystyle{IEEEtran}
\bibliography{bibliography}



\end{document}